\documentclass[conference]{IEEEtran}
\IEEEoverridecommandlockouts
\usepackage{cite}
\usepackage{amsmath,amssymb,amsfonts}
\usepackage{algorithmic}
\usepackage{graphicx}
\usepackage{subcaption}
\usepackage{graphicx}
\usepackage{textcomp}
\usepackage{xcolor}
\usepackage{multirow}
\usepackage{url}
\usepackage{algorithm}
\usepackage{algorithmic}
\def\BibTeX{{\rm B\kern-.05em{\sc i\kern-.025em b}\kern-.08em
    T\kern-.1667em\lower.7ex\hbox{E}\kern-.125emX}}
\begin{document}
\title{NEEDL-Bench: Dataset for Swiss Needle Cast and Stomata Detection in Microscopy Images\\
\thanks{We would like to thank the ADM Innovation Fund, Genome British Columbia, Canadian Forest Service, Natural Resources Canada and the Natural Sciences and Engineering Research Council of Canada for supporting this research. Needle sampling was supported by a GeneSolve project funded by Genome British Columbia.
}
}
\author{\IEEEauthorblockN{1\textsuperscript{st} Benjamin Blake}
\IEEEauthorblockA{\textit{Electrical and Computer Engineering} \\
\textit{University of Victoria}\\
Victoria, Canada \\
benjamincblake@uvic.ca}
\and
\IEEEauthorblockN{2\textsuperscript{nd} Declan McIntosh}
\IEEEauthorblockA{\textit{Electrical and Computer Engineering} \\
\textit{University of Victoria}\\
Victoria, Canada \\
declanmcintosh@uvic.ca}
\and
\IEEEauthorblockN{3\textsuperscript{rd} Jürgen Ehlting}
\IEEEauthorblockA{\textit{Department of Biology and Centre}\\
\textit{ for Forest Biology}\\
\textit{University of Victoria}\\
Victoria, Canada \\
je@uvic.ca}
\and
\IEEEauthorblockN{4\textsuperscript{th} Nicolas Feau}
\IEEEauthorblockA{\textit{Pacific Forestry Centre} \\
\textit{Canadian Forest Service, Natural }\\
\textit{Resources Canada}\\
Victoria, Canada \\
nicolas.feau@nrcan-rncan.gc.ca}
\and
\IEEEauthorblockN{5\textsuperscript{th} Joey B. Tanney}
\IEEEauthorblockA{\textit{Pacific Forestry Centre} \\
\textit{Canadian Forest Service, Natural }\\
\textit{Resources Canada}\\
Victoria, Canada \\
joey.tanney@nrcan-rncan.gc.ca}
\and
\IEEEauthorblockN{6\textsuperscript{th} Alexandra Branzan Albu}
\IEEEauthorblockA{\textit{Electrical and Computer Engineering} \\
\textit{University of Victoria}\\
Victoria, Canada \\
aalbu@uvic.ca}
}
\maketitle
\begin{abstract}
We present NEEDL-Bench, a microscopy detection benchmark for Swiss Needle Cast (SNC), a fungal disease of Douglas-fir trees. Douglas-fir is a keystone species of major ecological and economic importance as a softwood timber resource, and SNC affects productivity by forming sexual reproductive structures (pseudothecia) that emerge through the gas exchange pores (stomata) of the needles, thereby blocking gas exchange and compromising needle function. To date, there is no dataset for automatic computer vision detection of these structures, despite computer vision being well poised to standardize and viably scale severity measurements.
To address this, we present NEEDL-Bench, a dataset of 3250 annotated images from 1082 Douglas-fir needles, annotated for both keypoints and bounding-box detectors.
This dataset exhibits a challenging collection of features, including blur, poor object contrast, small objects of interest, and occlusions.
To better capture both the nominal distribution of the data and the full breadth of rare structures, we present two distinct evaluation splits: either random sampling from the collected images or sequential sampling to maximize structural diversity.
We evaluate multiple popular keypoint and bounding box methods for detection on this dataset as a baseline and observe a maximum F1 score of 0.8479, suggesting significant potential for gains from future development on this problem.
Further, we find that larger models generally do not show commensurate gains in performance on this dataset, indicating that improvements on this problem will not come from scaling laws but rather from domain-specific inductive biases.
\end{abstract}
\section{Introduction}

Douglas-fir (\textit{Pseudotsuga menziesii}) is a keystone ecological species of high commercial importance in both its native and introduced ranges. Within its native range, as well as in regions where it has been introduced as a plantation species (e.g., New Zealand and Europe), productivity can be threatened by Swiss needle cast (SNC). SNC is a foliar disease caused by the fungus \textit{Nothophaeocryptopus gaeumannii}, which can result in volume growth losses of up to 50\% in severe cases \cite{maguire2011ten}. The fungus forms its sexual reproductive structures (pseudothecia) through the gas exchange pores (stomata) of the needles, effectively occluding them; when a threshold level of stomata occlusion is reached, gas exchange is impaired, leading to chlorosis and premature needle loss \cite{maguire2011ten, stone2008predicting}. A changing climate marked by warmer winters and increased spring precipitation is expanding the disease's range and severity, while traditional disease control methods, such as fungicides, are ineffective or impractical at the landscape scale. Consequently, new biological control approaches are being explored alongside tree improvement programs aimed at identifying genetic tolerance to the disease \cite{graham2025vitro, montwe2021swiss}. Because future forest resilience depends on the rapid identification of tolerant genotypes, this requires extensive disease phenotyping of breeding populations; however, current assessment methods are time-consuming and not easily scalable.
\begin{figure}[!h]
    \centering
    \includegraphics[width=1.0\linewidth]{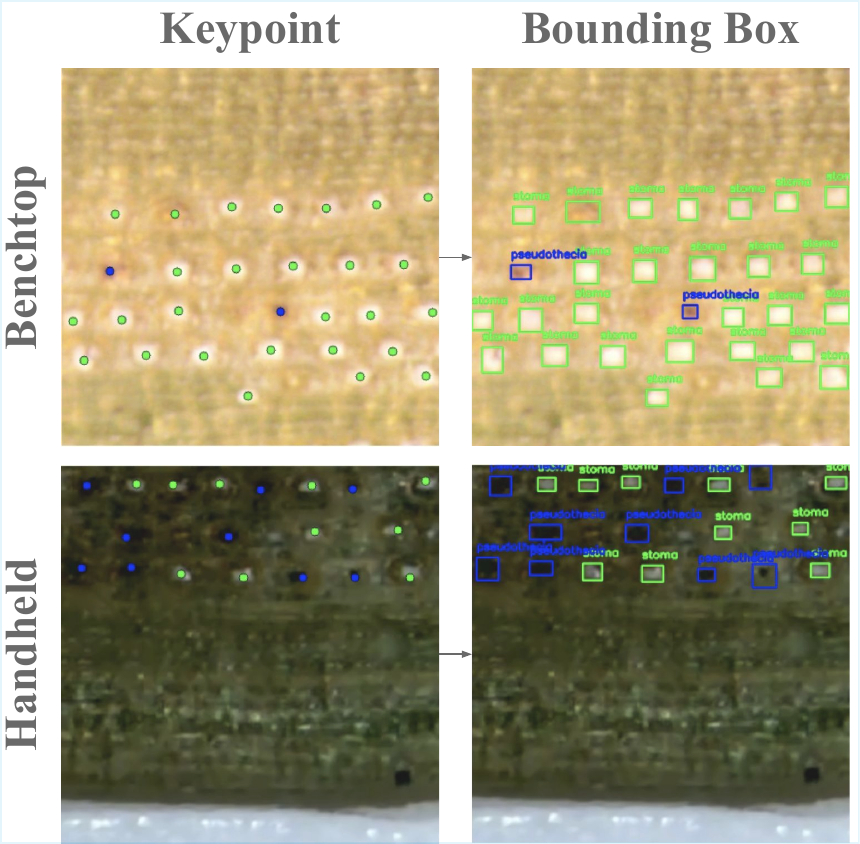}
    \caption{Example tiles from the NEEDL-Bench dataset, captured on both the benchtop and handheld microscopes. Both types of annotations are shown for the given images, keypoints and bounding boxes. Stomata are labelled in green, pseudothecia occlusions are labelled in blue.}
    \label{fig:intro}
\end{figure}
Currently, assessment of disease incidence and severity is done through visual inspection, either by manually counting the percentage of stomata occlusion on an individual needle or by visually scoring needle loss/retention at the branch or tree level, respectively \cite{winton2003comparison}. This process is arduous, resource-intensive, error-prone, and lacks scalability. To effectively cultivate resistant genotypes of Douglas-fir, forest pathologists and geneticists require large, accurate datasets of tree tolerance to serve as the foundation for genetic assessment \cite{winton2003comparison}. There is currently no image dataset focusing on SNC severity at the needle level. This dataset would enable computer vision scientists to offer a robust alternative solution to the assessment of the disease at a large scale.
\begin{figure*}
    \centering
    \includegraphics[width=1\linewidth]{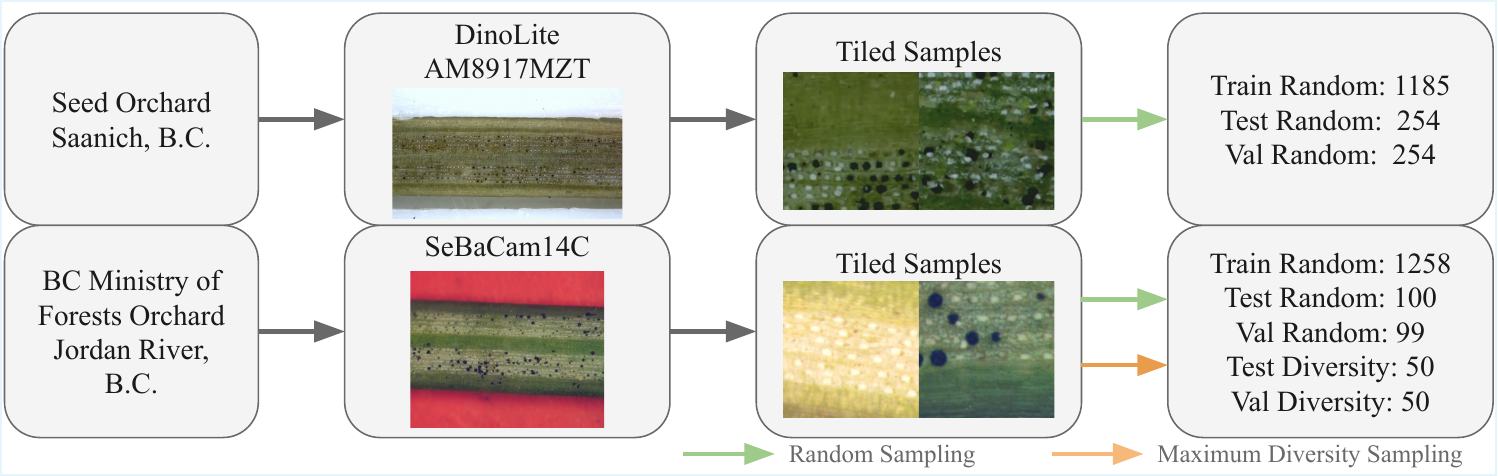}
    \caption{Summary of data collection pipeline from source trees, image capture, tile sampling, annotation, and data splits.}
    \label{fig:placeholder}
\end{figure*}
To broaden the range of applicable computer vision methods, we provide annotations for keypoint-level counting of SNC-occluded and clear stomata, along with corresponding minimum area bounding-box annotations. This allows for a direct comparison of keypoint and bounding-box detection methods for the counting and localization of these structures of interest. The needle images are sourced from two common microscope setups: one benchtop and the other handheld, with different apparatuses. This was done to ensure that the dataset represents a range of image collection methods possibly used in laboratories studying SNC \cite{winton2003comparison}.
This dataset of needle-level images poses a non-trivial challenge for existing computer vision methods. Issues arise, for instance, from the use of microscopy, where the depth of field is often very limited, leading to regions of the three-dimensional needles being out of focus. This lack of focus can be further exacerbated by low contrast between the stomata of interest and the needle substrate. Further, our dataset includes visually similar but distinct diseases, as well as foreign objects from needle sampling. A subset of samples were also stored in a freezer at -20 °C for a significant period of time before being photographed. These challenges lead to middling results from the selected model baselines, which are not resolved by increasing the model's scale. In summary, strong detection performance on the NEEDL-Bench dataset will need domain-specific inductive biases. We provide the NEEDL-Bench dataset\footnote{Dataset and code: \url{https://github.com/benjamincblake/NEEDL-BENCH-Dataset}} to support future work in developing and evaluating such approaches with the following important traits:
\begin{itemize}
    \item Large annotation scope of 3250 images, with annotations in centroid keypoint and minimum area bounding box formats.
    \item Evaluation splits that represent the true distribution of sampled data (Random Set) as well as the full diversity of rare sampled structures (Diversity Set).
    \item Challenging microscopy images, with selected benchmarks scoring at most an average F1 score of 0.8479.
    \item An empirical demonstration that representational scaling does not improve detection performance on this microscopy task, motivating future work on domain-specific inductive biases rather than larger general-purpose backbones.
\end{itemize}
The remainder of this paper is organized as follows: Section~\ref{sec:related} reviews related work in plant bioimaging and detection architectures relevant to small-object microscopy. Section~\ref{sec:dataset} describes the NEEDL-Bench dataset, including acquisition, preprocessing, annotation, and our two complementary sampling protocols. Section~\ref{sec:eval} details the evaluation methodology, metric definitions, and benchmark models. Section~\ref{Results} reports and analyzes results across acquisition modalities, sampling strategies, and detection paradigms. Section~\ref{sec:conclusion} concludes with a discussion of inductive-bias gaps, current limitations, and future directions.
\section{Related Works}
\label{sec:related}
To address the scarcity of large-scale annotated data in plant bioimaging, recent literature has established several foundational benchmarks that validate computer vision strategies for small, specialized microscopy datasets. For epidermal phenotyping, the LeafNet dataset \cite{li2022leafnet} provides ground truth for bright-field microscopy, demonstrating that hierarchical models—which first detect stomata to isolate pavement cells—can effectively handle noise and complex boundaries even with limited training data. This is complemented by the Hardwood and Populus datasets \cite{wang2024labeled}, which offer taxonomically diverse annotations for more than 10,000 images, facilitating the training of versatile detectors that generalize across species. In the domain of root microscopy, the PHDFM dataset \cite{wanner2024nfroot} serves as a key resource for confocal imaging, enabling the precise segmentation of developmental zones and apoplastic pH analysis in \textit{Arabidopsis} through supervised learning on 2D tissue slices. Additionally, RootNav 2.0 \cite{yasrab2019rootnav} validates the use of transfer learning for root architecture detection, showing that models pre-trained on large datasets of a single species can be adapted to smaller datasets with minimal fine-tuning, a crucial strategy for data-constrained research.
We consider three distinct and highly relevant architectures that capture the current trends for object detection and keypoint detection in computer vision: the transformer-based RF-DETR (Base) \cite{robinson2025rfdetr}, and the CNN-based YOLOv11 (Nano) in both its detection and pose estimation configurations \cite{jocher2024yolo11, maji2022yolo}. RF-DETR utilizes a Vision Transformer backbone (DINOv2 \cite{oquab2023dinov2}) with self-attention mechanisms to capture global context; however, unlike Convolutional Neural Networks (CNNs), transformers lack the inherent inductive biases of translation invariance and locality of image features \cite{dosovitskiy2021image}. Consequently, they typically require significantly larger datasets to learn these spatial relationships from scratch, potentially predisposing them to overfitting or slower convergence in limited-data regimes such as those in the presented dataset \cite{dosovitskiy2021image}. In contrast, the YOLOv11 architectures leverage convolutional priors that are often more sample-efficient for small datasets \cite{jocher2024yolo11}. Furthermore, we distinguish between the fundamental prediction tasks imposed by these models: while the standard RF-DETR and YOLOv11 detectors are tasked with resolving object boundaries (bounding boxes), a challenge in dense tissue where cell walls are often shared or ambiguous, the YOLOv11 Pose model reformulates the problem as keypoint centroid detection \cite{jocher2024yolo11}. This shifts the objective from delineating spatial extent to localizing object centers, a distinction that may prove critical if the visual ambiguity of boundaries outweighs the difficulty of identifying centroids.
\section{Dataset}
\label{sec:dataset}
\subsection{Data Collection}
The dataset comprises Douglas-fir needle samples collected from multiple sites on Vancouver Island, British Columbia, Canada, selected to capture a range of SNC severity and host maturity. To ensure the dataset addresses the practical constraints of forestry pathology, we employed two imaging workflows: benchtop and handheld, capturing the more controlled conditions of a benchtop microscope and the more portable, in-situ--compatible setup of a handheld microscope, respectively.
The collection sites for benchtop imaging were three plantations of a General Combining Ability population from the BC Ministry of Forests Douglas-fir breeding program near Jordan River, Lake Cowichan, and Campbell River, B.C. From these locations, needles were harvested from $>20$-year-old Douglas-fir trees exhibiting varying degrees of chlorosis and defoliation associated with SNC \cite{maguire2011swiss}. These samples were transported to the laboratory and stored at -20 °C until used for mounting and high-resolution imaging. In total, 907 needles were imaged using a standard benchtop microscopy setup, prioritizing high-magnification clarity and resolution and stable lighting conditions to serve as a high-quality baseline for the dataset.
To capture the variability introduced by field acquisition, a second cohort of samples was collected in Saanichton, B.C. Additional samples were taken from older Douglas-firs in the vicinity of the University of Victoria. Unlike the benchtop samples, these needles were imaged using a handheld Dino-Lite AM8917MZT microscope. This imaging setup produced more consistent images, as needles were constrained in a 3D-printed jig and stacks of images were taken at varying focal depths for each needle, then automatically stitched together to create a single in-focus image. We collected and imaged 175 needles using this handheld method.
\begin{figure*}[!h]
    \centering
    \begin{subfigure}[b]{0.19\textwidth}
        \centering
        \includegraphics[width=\textwidth]{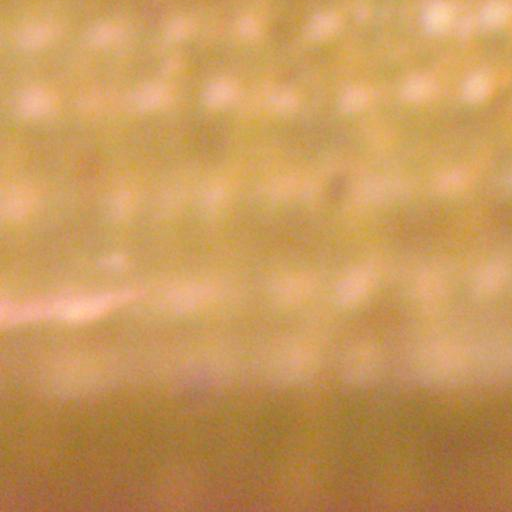}
        \caption{Blurring}
        \label{fig:diff1}
    \end{subfigure}
    \hfill
    \begin{subfigure}[b]{0.19\textwidth}
        \centering
        \includegraphics[width=\textwidth]{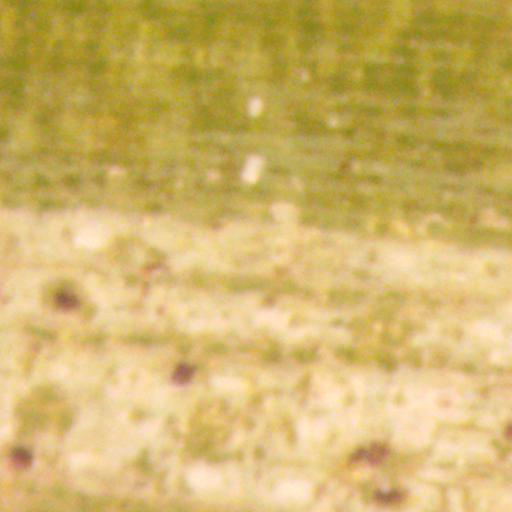}
        \caption{Poor Contrast}
        \label{fig:diff2}
    \end{subfigure}
    \hfill
    \begin{subfigure}[b]{0.19\textwidth}
        \centering
        \includegraphics[width=\textwidth]{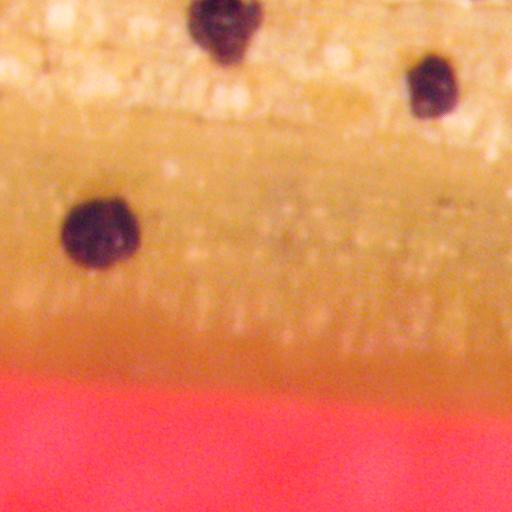}
        \caption{Displaced Pseudothecia}
        \label{fig:diff3}
    \end{subfigure}
    \hfill
    \begin{subfigure}[b]{0.19\textwidth}
        \centering
        \includegraphics[width=\textwidth]{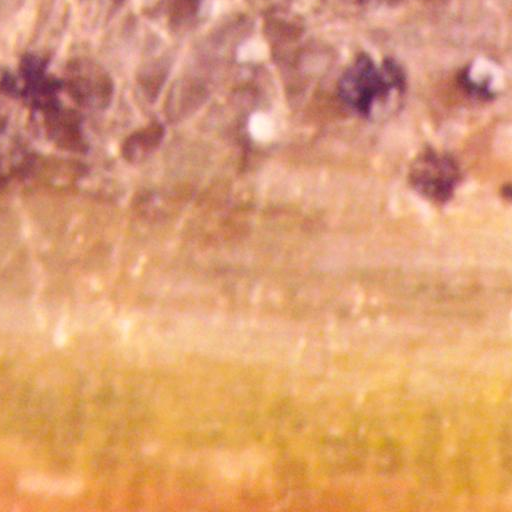}
        \caption{Partial Occlusion}
        \label{fig:diff4}
    \end{subfigure}
    \hfill
    \begin{subfigure}[b]{0.19\textwidth}
        \centering
        \includegraphics[width=\textwidth]{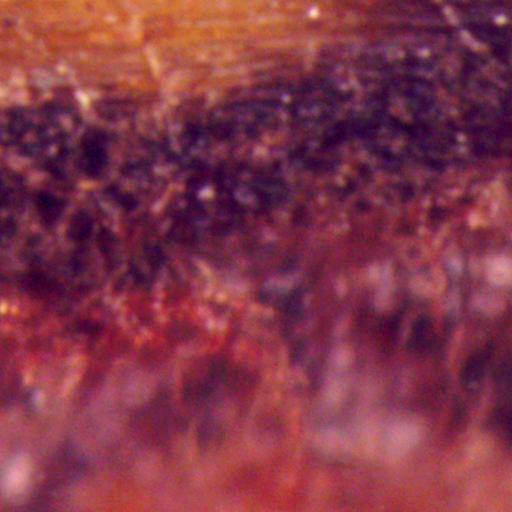}
        \caption{Full Occlusion}
        \label{fig:diff5}
    \end{subfigure}

    \caption{Qualitative examples of common challenges in the NEEDL-Bench dataset. Challenge figures are listed in order of prevalence in the dataset, with Blurring being the most common and Full Occlusion the least common of the challenges presented.}
    \label{fig:five_images}
\end{figure*}
\begin{table}[h]
    \centering
    \caption{Scale of needle images captured across imaging configurations.}
    \label{tab:Needle_Counts}
    \begin{tabular}{lc}
        \hline
        Source & Needles Captured \\
        \hline
        Benchtop & 907 \\
        Handheld & 175 \\
        \hline
    \end{tabular}
\end{table}
\subsection{Data Pre-Processing}
To ensure consistent scale across imaging methods, we scale the raw images captured by either the benchtop or handheld microscope to a spatial resolution of 360 pixels/mm. Once on a consistent scale, we split raw images into 256$\times$256 tiles. This is done for two reasons: first, to ensure the image aspect ratio in the dataset is consistent across imaging methods with disparate aspect ratios. Second to increase the relative size of the objects of interest compared to the tile size. This increase in relative scale prevents a known issue with existing object detectors, which perform poorly on small objects in given images \cite{mirzaei2023small}. All other preprocessing, such as normalization, was performed in accordance with the respective recommended practices for the methods evaluated in Section~\ref{Results}.
\subsection{Data Splits}
\label{sec:data splits}
The vast majority of needle samples exhibited only common expected structures with a long tail of rare, sometimes single-occurrence, needle conditions or co-located structures. We employed a dual-strategy sampling approach to curate a representative and challenging dataset from the set of tiles. First, we generated a randomly sampled subset to approximate the natural distribution of visual features for each of our image capture configurations. This sampling approximates the expected real-world distribution of features from captured needles. Second, to ensure our models were exposed to rare edge cases observed in the long tail of samples, we generated a maximum-diversity subset. Maximum diversity sampling was used to create a set of tiles whose distribution is uniform over feature space, in contrast to the uniformly random sampling over tiles. This combination creates evaluation sets which are either statistically representative of average infection rates and visual structures or broadly representative of all visual structures sampled.
To perform the maximum diversity sampling, we first describe each tile using a pre-trained feature extractor, specifically the flattened last feature layer of PyTorch's published ResNet50 weights, trained on ImageNet \cite{imagenet}. To reduce computational costs, we then project this feature space down to a $\mathbb{R}^{256}$ feature space by a fixed random linear projection. The Johnson-Lindenstrauss lemma bounds the maximum distortion introduced by this projection \cite{linear_pojection_limit}. We then randomly sample the first selected tile from the set. We then select the tile whose feature vector is furthest from the set of tiles selected so far and add it to our set. This continues until the desired number of tiles is selected. This is effectively a farthest-point sampling over the projected image feature space, similar to K-means++ sampling \cite{farthest, kmeans++}.
We use uniform random sampling for all data used in the training set to prevent biasing the training towards rare cases, and include a mixture of both the maximum-diversity and randomly sampled tiles in both the testing and validation splits. The benefit of the maximum diversity set being uniformly distributed across possible tile features is that it ensures the model is tested on the widest range of tile features possible for the collected data. For instance, in rare cases, the needle is fully occluded by disease growth, which is a novel texture compared to the rest of the dataset, but one still desirable to ensure a trained model generalizes to. We generate the maximum-diversity sets from only the needles captured in the larger benchtop Jordan River set, as its larger sampling scale contained more rare structures. Sets from different sources were kept separate to enable comparisons across sources in Section~\ref{Results}. Since maximum diversity sampling is a sequential process, we first take the first 50 samples for the testing set, then the subsequent 50 for the validation set. This means the model is tested on the ``hardest'' 50 samples due to being the most diverse.
A summary of the dataset splits, their sources, and sampling methods is shown in Table~\ref{tab:data_splits}. Maximum diversity sampling was performed only on the Benchtop set, as it had a larger number of captured needles to source from and therefore greater diversity.
\begin{table}[h]
    \centering
    \caption{Magnitude of data splits for NEEDL-Bench dataset.}
    \label{tab:data_splits}
    \begin{tabular}{l|l|rrr}
        \hline
        Source & Sampling Method & Train  & Validation  & Test  \\
        \hline
        Benchtop & Random            & 1258 & 99  & 100 \\
        Benchtop & Maximum Diversity & 0    & 50  & 50  \\
        Handheld & Random            & 1185 & 254 & 254 \\
        Handheld & Maximum Diversity & 0    & 0   & 0   \\
        \hline
                 & Total             & 2443 & 403 & 404 \\
        \hline
    \end{tabular}
\end{table}
\subsection{Automatic Conversion of Keypoints to Bounding Boxes}
\begin{figure}[!h]
    \centering
    \includegraphics[width=1.0\linewidth]{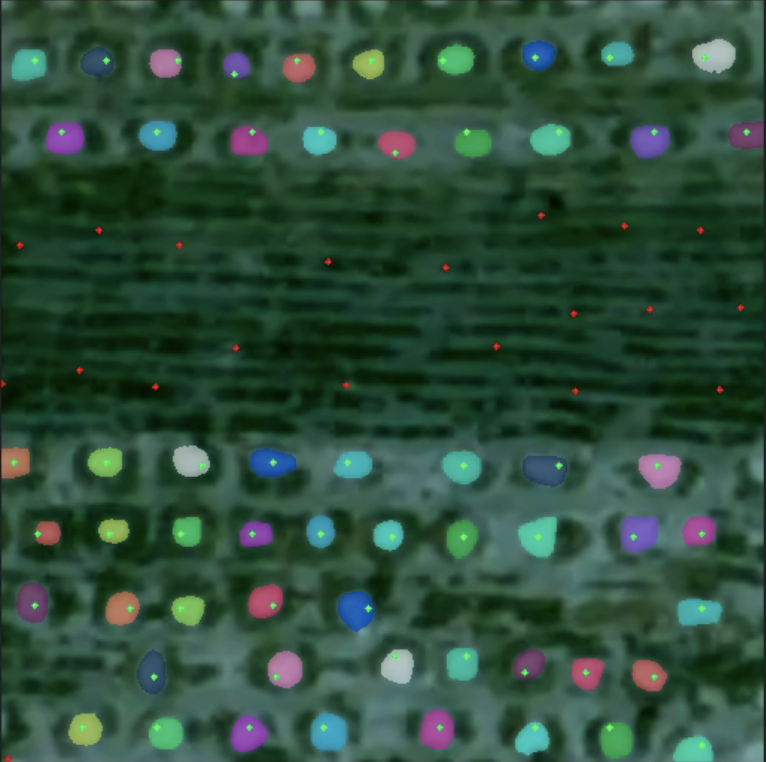}
    \caption{Visualization of automatic conversion of manually annotated keypoints to semantic segmentations using SAM2. Positive manual annotations are shown in green dots, automatically generated negative annotations are shown in red dots, and generated segmentations are displayed in varied colours.}
    \label{fig:sam_results}
\end{figure}
\begin{figure*}[htbp]
    \centering
    \makebox[\textwidth][c]{%
        \begin{subfigure}[b]{0.33\textwidth}
            \centering
            \includegraphics[width=\textwidth]{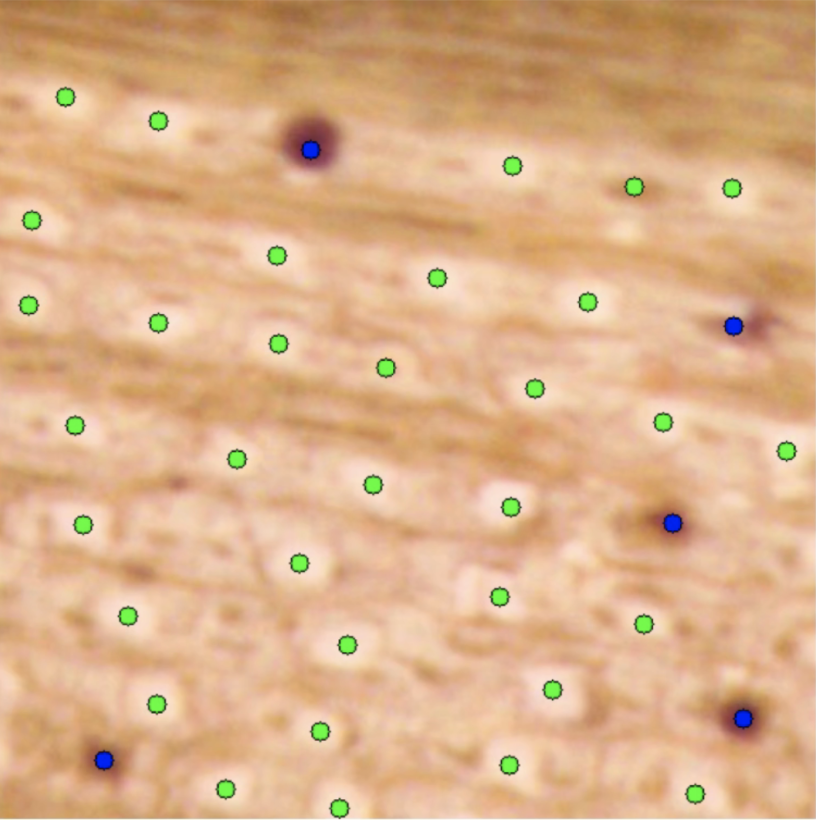}
            \caption{Keypoint Annotations}
            \label{fig:annot1}
        \end{subfigure}
        \hspace{0.01\textwidth}
        \begin{subfigure}[b]{0.33\textwidth}
            \centering
            \includegraphics[width=\textwidth]{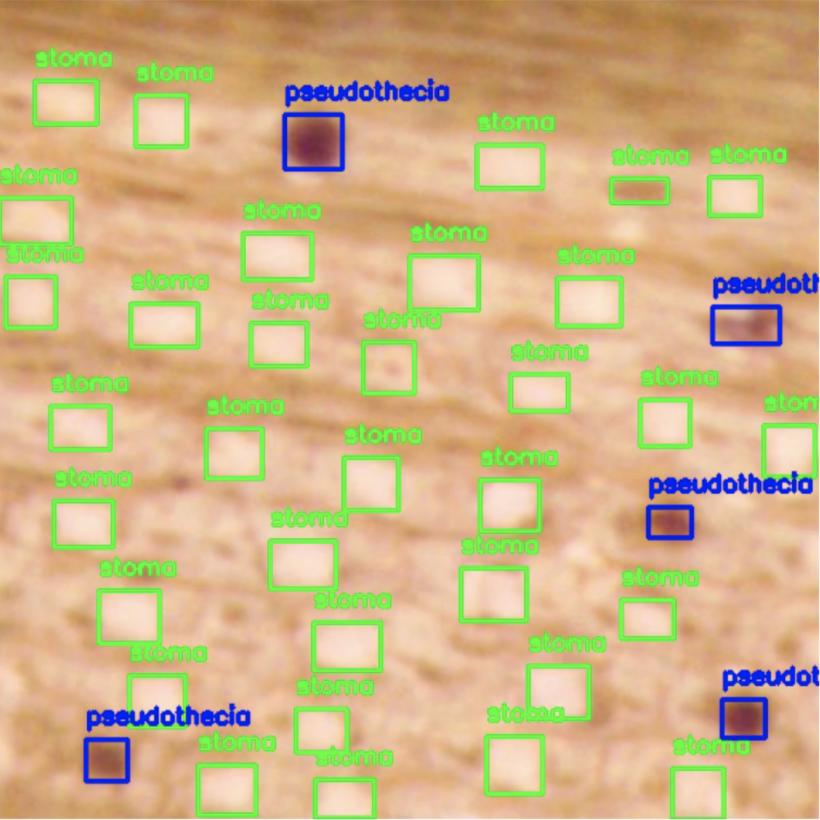}
            \caption{Bounding Box Annotations}
            \label{fig:annot2}
        \end{subfigure}
    }
    \caption{Visual comparison of dataset annotation styles. (a) illustrates the point-based annotations used for the keypoint dataset, while (b) displays the bounding box format used for YOLO training.}
    \label{fig:annotation_comparison}
\end{figure*}
The manually generated keypoint annotations were converted to bounding boxes semi-automatically using SAM2, a large zero-shot segmentation model \cite{ravi2024sam2}. SAM2 was seeded with manually created keypoint annotations, as well as randomly distributed negative keypoints at least 0.14 mm or 50 pixels from positive keypoints \cite{ravi2024sam2}. An example output instance segmentation from this method can be seen in Figure~\ref{fig:sam_results}. Then, minimum area bounding boxes were created around these segmentations. Segmentations that did not correspond to seeded keypoints were discarded. Segmentations that were either much smaller or much larger than expected for stomata or pseudothecia were assigned a default bounding box centred on the original annotation keypoint. This conversion's outputs were manually reviewed and found acceptable without requiring human manual edits. All keypoint annotations were produced by a single trained annotator with periodic spot-checks by a domain expert; we did not collect formal inter-annotator agreement statistics, and quantifying labeling noise via dual annotation of a held-out subset is left to future work.
\section{Evaluation Methodology}
\label{sec:eval}
\subsection{Evaluation Metrics}
SNC-occluded and non-occluded stomata evaluation for the measurement of severity requires detection and localization for counting objects of interest \cite{manter2000pseudothecia}. We follow standard practice for matching predictions to ground truth, then derive precision, recall, and F1 from the resulting matches.

\paragraph{Bounding-box matching} Following the PASCAL VOC convention \cite{everingham2010pascal}, a predicted box $b_p$ is matched to a ground-truth box $b_g$ when their Intersection-over-Union
\begin{equation}
\mathrm{IoU}(b_p, b_g) = \frac{|b_p \cap b_g|}{|b_p \cup b_g|} \geq 0.5.
\end{equation}

\paragraph{Keypoint matching} Predicted and ground-truth keypoints are paired greedily by minimum Euclidean distance, after which any pair with separation
\begin{equation}
\lVert k_p - k_g \rVert_2 > \tau
\end{equation}
is rejected. We set $\tau = 36$ pixels ($\approx 0.1$ mm) at our 360 px/mm resolution---half of the empirically observed minimum inter-stomatal spacing of 0.2 mm---guaranteeing each ground-truth keypoint can match at most one prediction.

All reported scores reflect a single training run per (model, split) configuration; differences within $\approx 1$--$2$ F1 points should therefore be interpreted with caution.
\subsection{Benchmark Models}
We evaluate three distinct architectures on our dataset to establish a baseline: the object detection framework YOLOv11 (in both bounding box and pose-estimation configurations), the transformer-based RF-DETR, and a Fully Convolutional Network (FCN) \cite{long2015fully} based on ResNet50  \cite{he2016deep}. All results were generated on the same machine using a single RTX 5060 Ti.
We utilized the YOLOv11 architecture, specifically the Nano variant, which is a highly relevant architecture for lightweight, real-time object detection \cite{jocher2024yolo11}. We evaluated two variations of this architecture to isolate and evaluate the effects of different target tasks on detection and localization performance on our dataset:
\begin{itemize}
    \item \textbf{YOLOv11-Detect:} A standard bounding-box detection model.
    \item \textbf{YOLOv11-Pose:} A keypoint estimation model designed to predict specific structural points, modified to predict each stoma in isolation.
\end{itemize}
For a transformer-based baseline, we employed RF-DETR (Receptive Field DETR) in its base configuration  \cite{robinson2025rfdetr}. Unlike YOLO's pure CNN approach, RF-DETR leverages transformer encoders and decoders to model global context and object relationships directly using an attention mechanism \cite{robinson2025rfdetr}. \
Similarly, as a baseline for larger CNN-based keypoint detection, we implemented a custom model based on the FCN-ResNet50 architecture \cite{he2016deep}. We use torchvision's FCN-ResNet50, whose ResNet50 backbone is pre-trained on ImageNet \cite{imagenet} and whose FCN head is trained on a COCO subset using the 20 PASCAL VOC categories \cite{lin2014microsoft, everingham2010pascal}. We modified the classifier head to a $1 \times 1$ convolution, producing 2-channel keypoint predictions via non-maximal suppression. We used an online hard negative mining loss, evaluated separately for each category.
\subsubsection{Implementation and Training Configuration}
Both YOLO models were initialized with default pre-trained weights to leverage transfer learning.
\begin{itemize}
    \item \textbf{Detection Model:} Trained for a maximum of 150 epochs with early stopping and an image size of 640 and multi-scale training disabled.
    \item \textbf{Pose Model:} Trained for a maximum of 150 epochs with early stopping. Recognizing that the target objects are small points, we modified the standard augmentation pipeline to be less aggressive: mosaic augmentation probability was reduced to 0.5, and the box loss gain was lowered to 0.05 to de-emphasize bounding box accuracy in favour of keypoint accuracy (pose loss gain increased to 12.0). Bounding box information for the keypoints was automatically generated at a fixed size of 36$\times$36 px (matching the keypoint association radius $\tau$ from Section~\ref{sec:eval}), uncorrelated with the actual stomata or pseudothecia sizes, to force the model to consider only the keypoint annotations.
\end{itemize}
The RF-DETR model was trained using its default recipe without modification.
\begin{table*}[!htb]
    \centering
    \caption{Test set results of selected baselines across the entire NEEDL-Bench dataset. Precision, recall, and F1 scores are calculated at thresholds to maximize F1 on the validation set.}
    \label{tab:full_set_results}
    \begin{tabular}{ll|ccc|ccc|c}
        \hline
        & & \multicolumn{3}{c|}{Pseudothecia} & \multicolumn{3}{c|}{Stomata} & Average \\
        \cline{3-8}
        Split & Model & F1 & Precision & Recall & F1 & Precision & Recall & F1 \\
        \hline
        Full & FCN-ResNet & 0.8078 & 0.8007 & 0.8150 & 0.8794 & 0.8905 & 0.8685 & \textit{0.8436} \\
        Full & YOLOv11n-Pose & 0.7479 & 0.8225 & 0.6857 & 0.8859 & 0.8653 & 0.9075 & 0.8169 \\
        Full & RF-DETR & 0.7804 & 0.8662 & 0.7100 & 0.9153 & 0.9312 & 0.9000 & \textbf{0.8479} \\
        Full & YOLOv11n & 0.7868 & 0.8440 & 0.7368 & 0.9071 & 0.9206 & 0.8940 & \underline{0.8470} \\
        \hline
    \end{tabular}
\end{table*}
The FCN-ResNet model was trained under the following scheme:
Given that our detection problem configuration is closely adjacent to the field of pose estimation problems, we can apply similar methods for data processing and model training  \cite{maji2022yolo}. We followed standard procedure for pose estimation problems by convolving a Gaussian over impulse keypoint annotations to generate smoother targets that are easier for the model to learn, then using Non-Maximal Suppression techniques to recover keypoint predictions from the output heatmaps \cite{belagiannis2017recurrent}. Below are the specifics of our training recipe used to generate our results.
\begin{itemize}
    \item \textbf{Data Preparation:} Input images were the standard $256\times256$ tiles generated during pre-processing. We applied standard ImageNet normalization to all inputs to align with the pre-trained ResNet50 backbone. No additional geometric augmentations were applied during the training loop.
    \item \textbf{Loss Function:} We implemented a custom Hybrid Spatial Online Hard Negative Mining (OHNM) loss, an OHEM-style \cite{shrivastava2016training} variant restricted to negative pixels, tailored for small object segmentation under class imbalance. This loss combines three components: (1) a pixel-wise Mean Squared Error (MSE) calculated on foreground pixels in a neighbourhood around keypoint annotations; (2) a false-positive penalty of the MSE of the complement of the foreground classes  $\lambda=2.5$; and (3) an OHNM term that computes the loss on the top $2\%$ of pixels with the highest error rates per batch to focus learning on difficult regions.

    \item \textbf{Learning Rate and Optimizer:} The model was optimized using the Adam optimizer with a fixed learning rate of $1\times10^{-4}$  \cite{kingma2015adam}. Training proceeded for a maximum of 500 epochs without a learning rate decay schedule. To prevent overfitting, the model weights that maximized the sum of the Pseudothecia and Stomata AUROC scores on the validation set were saved.

\end{itemize}
\begin{table*}[!htb]
    \centering
    \caption{Results of selected baselines broken down to scores on each split of the NEEDL-Bench test set. Precision, recall, and F1 scores are calculated at thresholds to maximize F1 on the validation set.}
    \label{tab:per_split_results}
    \begin{tabular}{ll|ccc|ccc|c}
        \hline
        & & \multicolumn{3}{c|}{Pseudothecia} & \multicolumn{3}{c|}{Stomata} & Average \\
        \cline{3-8}
        Split & Model & F1 & Precision & Recall & F1 & Precision & Recall & F1 \\
        \hline
        Handheld Random & FCN-ResNet & 0.8052 & 0.7820 & 0.8298 & 0.8750 & 0.8921 & 0.8585 & 0.8401 \\
        Handheld Random & YOLOv11n-Pose & 0.7379 & 0.8138 & 0.6750 & 0.8772 & 0.8514 & 0.9048 & 0.8076 \\
        Handheld Random & RF-DETR & 0.7916 & 0.8509 & 0.7400 & 0.9184 & 0.9375 & 0.9000 & 0.8550 \\
        Handheld Random & YOLOv11n & 0.8027 & 0.8670 & 0.7474 & 0.9114 & 0.9284 & 0.8951 & \underline{\textbf{\textit{0.8571}}} \\
        \hline
        Maximum Diversity & FCN-ResNet & 0.8649 & 0.8485 & 0.8819 & 0.9107 & 0.8853 & 0.9377 & \underline{\textbf{\textit{0.8878}}} \\
        Maximum Diversity & YOLOv11n-Pose & 0.8214 & 0.8419 & 0.8018 & 0.9300 & 0.9205 & 0.9397 & 0.8757 \\
        Maximum Diversity & RF-DETR & 0.7526 & 0.7882 & 0.7200 & 0.9250 & 0.9405 & 0.9100 & 0.8388 \\
        Maximum Diversity & YOLOv11n & 0.7404 & 0.8162 & 0.6774 & 0.9202 & 0.9434 & 0.8982 & 0.8303 \\
        \hline
        Benchtop Random & FCN-ResNet & 0.8155 & 0.9265 & 0.7283 & 0.8669 & 0.8656 & 0.8682 & \underline{\textbf{\textit{0.8412}}} \\
        Benchtop Random & YOLOv11n-Pose & 0.7806 & 0.8756 & 0.7041 & 0.8949 & 0.9083 & 0.8819 & 0.8378 \\
        Benchtop Random & RF-DETR & 0.7232 & 0.9342 & 0.5900 & 0.8745 & 0.9241 & 0.8300 & 0.7989 \\
        Benchtop Random & YOLOv11n & 0.7330 & 0.8700 & 0.6333 & 0.8684 & 0.8906 & 0.8472 & 0.8007 \\
        \hline
    \end{tabular}
\end{table*}
\section{Results}
\label{Results}
We present an evaluation of benchmark models in Table \ref{tab:full_set_results}. The results are analyzed to benchmark baseline model performance on the NEEDL-Bench dataset and to characterize the dataset's unique challenges, including acquisition modalities (Handheld vs. Benchtop) and sampling strategies (Random vs. Maximum Diversity). We also present results on the NEEDL-Bench dataset, split-per-split, in Table \ref{tab:per_split_results} to evaluate across data acquisition methods and sources, as well as under maximum diversity and random sampling.
\subsection{Architectural Inductive Bias vs. Representational Power}
An examination of model capacity, measured by the number of trainable parameters and performance on the NEEDL-Bench dataset, reveals that increasing representational power does not garner superior results on this microscopy task. The FCN-ResNet model, with 35.3M parameters, achieves an average F1 score of 0.8436 on the full split. In contrast, the YOLOv11n architecture, with only 2.6M parameters, achieves a comparable F1 of 0.8470. The RF-DETR model (29.0M parameters) performs similarly, with a score of 0.8479. Full comparisons of model sizes can be seen in Table~\ref{tab:model_comparison}.
This lack of significant improvement with scaling trainable parameters suggests that performance on this dataset is not bounded by raw capacity and is more likely to be improved by task-specific inductive biases rather than by scaling laws. The lightweight CNN architectures (YOLOv11), which possess strong translation invariance and locality biases, appear better suited for the repetitive, dense cellular structures of the needle tissue than the heavier, global-context focused Transformer models, which generally require larger datasets to converge \cite{dosovitskiy2021image}.
\subsection{Task-Specific Performance: Pose Estimation vs. Bounding Box}
We observe a distinct performance trade-off between pose estimation (keypoint) and object detection (bounding box) formulations, specific to the test split considered:
\begin{itemize}
    \item \textbf{Random Sampling:} In the Handheld Random and Benchtop Random splits, bounding box detection models generally outperform or match pose estimation models. For example, in the Handheld Random split, YOLOv11n (Detect) achieves an average F1 of 0.8571, surpassing YOLOv11n-Pose at 0.8076.
    \item \textbf{Maximum Diversity:} Conversely, for the Maximum Diversity split (which targets rare textures and edge cases), pose estimation models demonstrate superior robustness. The FCN-ResNet and YOLOv11n-Pose achieve average F1 scores of 0.8878 and 0.8757, respectively, outperforming the RF-DETR (0.8388) and YOLOv11n-Detect (0.8303).
\end{itemize}
This inversion suggests that bounding box regressors struggle with the ambiguity of object edges in the ``hard'' diversity cases, which feature blurring and occlusions. Defining the spatial extent of a blurry stoma is inherently more difficult than localizing its centroid, making pose estimation a potentially more robust strategy for the dataset's diverse edge cases.
Notably, as bounding box models outperform on the Random sets, this could indicate that richer bounding box annotations, showing the entire area of interest and the boundaries of a given object, may perform better when deployed on more common structures.
\subsection{Class-Level Analysis and Detection Difficulty}
Across all models and splits, Pseudothecia detection consistently lags behind Stomata detection. For instance, in the Full split, the best-performing model (RF-DETR) achieves an F1 of 0.9312 for Stomata but only 0.8662 for Pseudothecia. Further, the Recall for Pseudothecia generally trails its Precision at the optimal F1 threshold. For the YOLOv11n-Pose model on the Full split, Pseudothecia Precision is 0.8225 while Recall is 0.6857. This indicates the presence of hard positive samples that are consistently missed by the models, likely due to specific challenges such as partial occlusion or displaced structures, rather than simple class imbalance. Examples of partial occlusion can be seen in the top-right pseudothecia annotation in Figure~\ref{fig:annotation_comparison}, where the stoma is only partially occluded by the pseudothecia.
\subsection{Analysis Across Acquisition Modality}
Contrary to the initial assumption that the laboratory setup would provide a ``high-quality baseline,'' the Benchtop images proved to be the more challenging subset. Comparing the ``Random'' splits in Table \ref{tab:per_split_results}, the images captured with the Benchtop setup resulted in an average test-time F1 score approximately 0.0203 lower than those captured with the Handheld jig (Average F1 across models: $\approx 0.82$ for Benchtop Random vs. $\approx 0.84$ for Handheld Random).
This performance gap reflects the more challenging conditions in the Benchtop dataset, including less controlled lighting, increased blur in single-image captures, and reduced needle quality from storage freeze–thaw cycles. By contrast, the Handheld dataset used a constrained needle-capture jig and multi-focus image stacking to generate fully in-focus images with reduced blur.
\subsection{Efficiency Analysis of Baselines}
\begin{table}[!h]
    \centering
    \caption{Comparison of architectures used for NEEDL-Bench dataset baselines. (*RF-DETR uses a hybrid architecture of a Convolutional  for feature extraction, then a Transformer for detection)}
    \label{tab:model_comparison}
    \setlength{\tabcolsep}{4pt}
    \begin{tabular}{l|llcc}
        \hline
        Model  & Architecture & Annotation & Params (M) & Speed (ms) \\
        \hline
        YOLOv11n          & CNN       & BBox       & 2.6  & 2.5  \\
        RF-DETR           & Hybrid*   & BBox       & 29.0 & 22.2 \\
        YOLOv11n-Pose     & CNN       & Kpts       & 2.9  & 2.1  \\
        FCN-ResNet        & CNN       & Kpts       & 35.3 & 5.4  \\
        \hline
    \end{tabular}
\end{table}
Table \ref{tab:model_comparison} presents a comparative analysis of model architecture, parameter count, and inference latency across the selected baselines. An examination of these metrics reveals an expected trade-off between architectural complexity (model size) and computational throughput (inference speed).
\begin{table}[!h]
    \centering
    \caption{Comparison of average bounding box results on the full test set for different sizes of the YOLO11 object detection model. Notably, increasing parameters does not increase performance on the problem.}
    \label{tab:yolo_scales}
    \begin{tabular}{l|cccc}
        \hline
        Model & Params (M) & F1 & Precision & Recall \\
        \hline
        YOLOn & 2.6  & 0.847 & 0.882 & 0.815 \\
        YOLOl & 25.3 & 0.832 & 0.858 & 0.810 \\
        YOLOx & 65.9 & 0.840 & 0.879 & 0.805 \\
        \hline
    \end{tabular}
\end{table}
As detailed in Table \ref{tab:model_comparison}, the Transformer-based RF-DETR architecture exhibits the lowest throughput and highest latency among the evaluated detection models. Despite a parameter count of 29.0M, comparable to that of FCN-ResNet50, it achieves an inference speed of 22.2 ms, nearly an order of magnitude slower than the CNN-based YOLO variants. This latency can be attributed to the $O(n^2)$ computational complexity of Transformer self-attention mechanisms, where n is the number of image patches \cite{liu2021swin}. Unlike the CNN backbones, which leverage inductive biases of translation invariance and feature locality to optimize processing, RF-DETR relies on learning global contexts \cite{robinson2025rfdetr}, resulting in a significantly heavier computational burden during inference without a commensurate gain in detection accuracy on this dataset.
In contrast, the YOLOv11 Nano architectures demonstrate superior efficiency, achieving results comparable to other baselines while running faster and requiring fewer model parameters. The YOLOv11n (Detect) model utilizes only 2.6M parameters and achieves an inference speed of 2.5 ms. Notably, the YOLOv11n-Pose variant proves to be the most efficient model in the benchmark; despite a slight increase in parameters to 2.9M to accommodate keypoint regression, it achieves the fastest inference time of 2.1 ms. This validates the ``Nano'' architectural design priority on computational efficiency and suggests that lightweight CNNs are well-suited for high-frequency processing of needle microscopy data. To verify this, we compare the nano model on the larger YOLOl and YOLOx architectures \cite{maji2022yolo}. The comparison of these models is seen in Table~\ref{tab:yolo_scales}. Given that the nano model largely outperforms its much larger counterparts, this reinforces our inference that further advancements in the detection of our problem do not stem from increasing model size, but instead from the inductive biases present in the model architecture.
Finally, the Custom FCN-ResNet50 represents the highest parameter overhead in the benchmark at 35.3M parameters. While it offers granular, pixel-wise localization, its reliance on the deep ResNet50 backbone creates a bottleneck, resulting in an inference speed of 5.4 ms. Although this is roughly $4\times$ faster than the Transformer-based RF-DETR, it is still more than double the latency of the YOLOv11n-Pose model. Consequently, while the FCN-ResNet50 remains a robust baseline for keypoint localization tasks where model size is not a constraint, it is less viable than YOLO architectures for applications requiring minimal storage and real-time inference, such as orchard field analysis.
\section{Conclusion}
\label{sec:conclusion}
This paper presents the NEEDL-Bench dataset, focused on detecting Douglas-fir stomata and the fungal sexual structures (pseudothecia) that emerge through them. SNC mitigation is of key interest for protecting the ecosystem services and economic value of Douglas-fir on the landscape, especially in the context of climate change adaptation. The NEEDL-Bench dataset comprises 3250 expert-annotated tiles from images of 1082 needles, collected with two distinct image-capture methods on Vancouver Island, B.C., Canada. NEEDL-Bench provides annotations for the detection and localization of these key structures across the entire dataset, including both keypoints on object centers and minimum-area bounding boxes, enabling future methods to consider either or both supervised signals. The NEEDL-Bench dataset also provides evaluation data splits for both randomly sampled and maximum-diversity samples, allowing evaluation of methods' performance on average nominal data as well as on the full diversity of structures observed during sampling.
Based on evaluations of multiple popular baselines in both keypoint and bounding-box object detection, it was found that there is significant room for improvement with this dataset, especially in detecting challenging pseudothecia. The best-performing method was the larger RF-DETR, with an average F1 score of 0.8479, but we did not find that performance scaled with larger models' representational power. This suggests that the largest gains are likely to be found in domain-specific inductive biases rather than scaling laws, making it an interesting problem for future methods. For this reason, we believe future research on this benchmark should focus on preprocessing and domain-specific model architecture considerations, especially for handling low-contrast, blurry images and ambiguous or subtle occlusions from pseudothecia.

\paragraph{Limitations and future work} Several aspects of NEEDL-Bench warrant further investigation. First, all keypoint annotations were produced by a single trained annotator with periodic spot-checks by a domain expert; quantifying labeling noise via dual annotation of a held-out subset is left to future work. Second, the maximum-diversity evaluation protocol is currently restricted to the benchtop modality, since the larger benchtop pool yielded greater structural diversity; expanding diversity sampling to the handheld modality would strengthen claims about diversity-set generalization.
\bibliographystyle{IEEEtran}
\bibliography{main}
\end{document}